\documentclass[10pt,twocolumn,letterpaper]{article}

\usepackage{cvpr}
\usepackage{times}
\usepackage{epsfig}
\usepackage{graphicx}
\usepackage{amsmath}
\usepackage{amssymb}
\usepackage{amsmath}
\usepackage{lipsum}
\usepackage{graphics}

\usepackage{color}
\definecolor{dkgreen}{rgb}{0,0.6,0}
\definecolor{gray}{rgb}{0.5,0.5,0.5}
\definecolor{mauve}{rgb}{0.58,0,0.82}

\cvprfinalcopy 

\def\cvprPaperID{****} 

\ifcvprfinal\fi
\begin{document}

\title{Review on 6D Object Pose Estimation with the focus on Indoor Scene Understanding}

\author{Negar Nejatishahidin\\
George Mason University\\
Fairfax, VA, USA\\
{\tt\small nnejatis@gmu.edu} 
\and
Pooya Fayyazsanavi\\
George Mason University\\
Fairfax, VA, USA\\
{\tt\small pfayyazs@gmu.edu}
}



\maketitle
\thispagestyle{empty}

\begin{abstract}
6D object pose estimation problem has been extensively studied in the field of Computer Vision and Robotics. It has wide range of applications such as robot manipulation, augmented reality, and 3D scene understanding. With the advent of Deep Learning, many breakthroughs have been made; however, approaches continue to struggle when they encounter unseen instances, new categories, or real-world challenges such as cluttered backgrounds and occlusions. In this study, we will explore the available methods based on input modality, problem formulation, and whether it is a category-level or instance-level approach. As a part of our discussion, we will focus on how 6D object pose estimation can be used for understanding 3D scenes.

\end{abstract}
\section{Introduction}
\begin{figure}[tbp]
\centerline{\includegraphics[width=\linewidth]{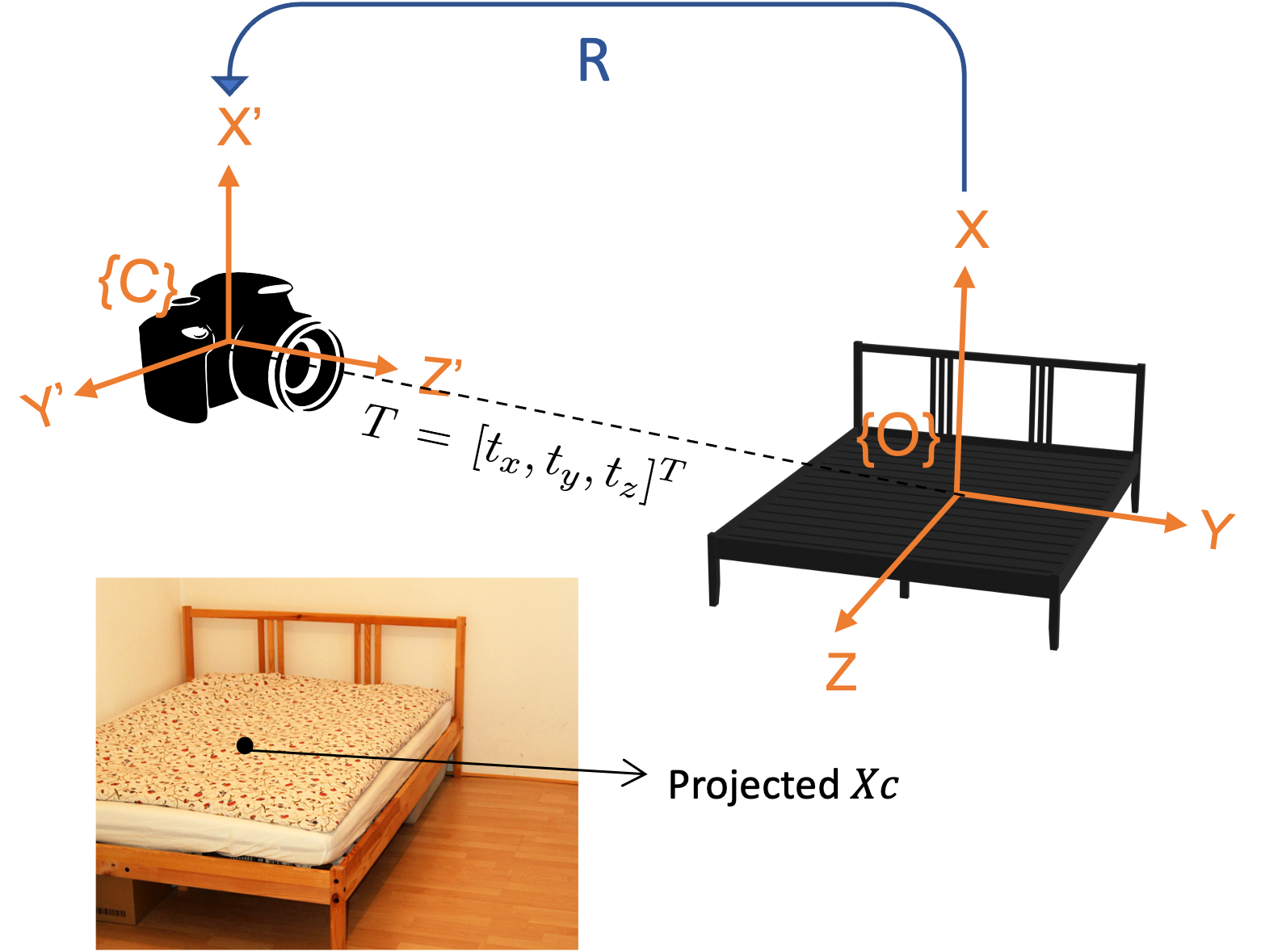}}
\caption{This figure illustrates the pose definition. $\{C\}$ is camera frame, $\{O\}$ is object frame, $R$ is the rotation matrix, $T$ is the translation.}
\label{coord}
\end{figure}

An object's 6D pose is defined as its 3D rotation and 3D translation in the camera coordinate frame. An illustration of the 6D pose concept is shown in Figure \ref{coord}. 
To reason about the importance of pose estimation, we can start with its application in Augmented Reality (AR) \cite{Marchand2016PoseEF}. In order to virtually place an object in an environment, information such as the 6D pose and scale of the objects in the scene can be helpful. Pose estimation models are commonly used in robot grasping as well \cite{Eppner2017LessonsFT}. The Amazon Picking Challenge \cite{Eppner2017LessonsFT} is one of the most well-known challenges that has heavily benefited from 6D pose estimation models. Furthermore, improving the performance of the object-oriented Simultaneous Localization and Mapping (SLAM) is directly influenced by camera-object constraints made with accurate pose estimation \cite{SLAM++, DBLP:journals/corr/Mur-ArtalMT15}. In addition, 3D detection of cars and motor vehicles \cite{Mousavian20173DBB, Geiger2013VisionMR} is essential to autonomous driving. Lastly, a holistic 3D understanding of all objects' poses, and geometry is required to transform an RGB image into a full 3D CAD representation \cite{DBLP:journals/corr/IzadiniaSS16, DBLP:journals/corr/abs-2002-12212, Dahnert2021Panoptic3S}. This representation can help to reduce the gap between the real world and the synthetic data.\\

To gain a better understanding of pose estimation and the pros and cons of the various methods, we have categorized the area according to its main components. 6D Pose estimation can be defined as either classification \cite{nejatishahidin2022object, pix3d, DBLP:journals/corr/abs-2109-08141, Poirson2016FastSS}, regression \cite{Wang2019NormalizedOC}, or classification followed by the regression task \cite{DBLP:journals/corr/MousavianAFK16}. It can also be divided based on the input modality. The RGB images \cite{Hu2020SingleStage6O,DBLP:journals/corr/PoirsonAFLKB16,Hodan2020EPOSE6} 
and RGB-D images \cite{ElGaaly2012RGBDOP,DeepSlidingShapes, 8237757} are the most common input modalities. A method either aims to solve the problem of pose estimation for instances \cite{ElGaaly2012RGBDOP, hodan2020bop,hodan2018bop, DBLP:journals/corr/abs-2008-08465}, or for categories \cite{pix3d,Georgakis_2018_CVPR, 8237757, Wang2019NormalizedOC, DBLP:journals/corr/abs-2011-12912}. In most instance-level approaches, the effectiveness of the method is evaluated for tabletop objects (e. g. a can, cup, bottle, or camera). In contrast, most of the Category level approaches focus on the main categories of objects (e.g. sofas, chairs, tables, desks, and cars). These approaches address the pose of the unseen instances within the seen category.

\begin{figure}[htbp]
\centerline{\includegraphics[width=\linewidth]{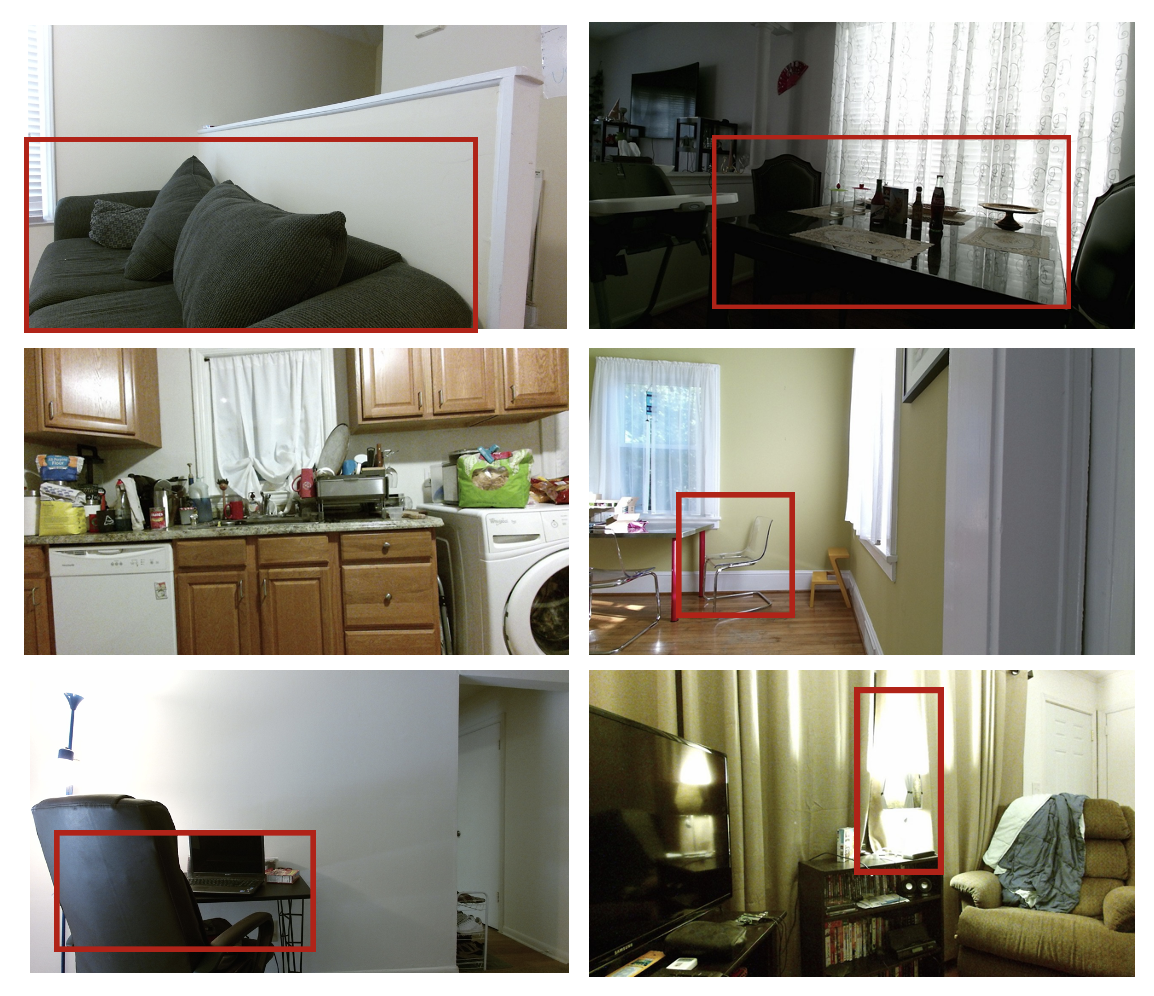}}
\caption{Real-world challenges from Active vision dataset~\cite{DBLP:journals/corr/AmmiratoPPKB17}; occlusion, truncation, cluttered background, large lighting variations, glassiness, and shiny objects. Such challenges impact the accuracy of pose estimation models..}
\label{challengess}
\end{figure}

The current approaches\cite{Hodan2018ASO,DBLP:journals/corr/MousavianAFK16} to pose estimation typically fail when faced with real-world challenges like cluttered backgrounds, occlusions, truncation, different lighting, dark objects, glossiness, and shiny objects. Figure \ref{challengess} shows some of the real-world challenges. 

Moreover, these methods cannot reliably handle unseen objects, significant differences in appearance between instances within the same category. The absence of 3D CAD models for all instances, and difficulty in understanding geometry structures are just some of the main reasons for these shortcomings. 


Most of the proposed approaches target rigid objects, however pose estimation of deformable or articulated objects has yet to be addressed \cite{6942687}. Glassy \cite{Phillips2016SeeingGF}, shiny, or reflective objects are also challenging for pose estimation. The majority of existing approaches for top-of-the-table objects are instance-level. More recent works \cite{Wang2019NormalizedOC} are addressing it at the level of categories, but still, they only consider categories with slight variation in appearance. Additionally, models that propose techniques for estimating the pose of indoor objects on real-world data using category-agnostic approaches \cite{nejatishahidin2022object, DBLP:journals/corr/abs-2106-01178} and usually do not perform well enough to be applied in practice. 

\section{What is 6D Pose?}
An object 6D pose is defined as a 3D rotation $R\in SO(3)$ and 3D translation $T=[t_x,t_y,t_z]^T$ matrix between each point $X_o= [x_o, y_o, z_o]^T$ of an object defined in the object frame $\{O\}$ and the same point $X_c= [x_c,y_c,z_c]^T$ defined in the camera frame $\{C\}$. For rigid objects all the points have the same rotation and translation matrix, transformation matrix. In general, pose parameterizes the orientation and translation of an object in camera coordinate frame. As illustrated in Figure \ref{coord}, the 3D translation $T = [x_t, y_t, z_t]^T$ can be interpreted as origin of the object coordinate frame in the camera frame, and the rotation 
$R = R_{z}(\alpha) R_{y}(\beta) R_{x}(\gamma)$ are the angles between each object frame axis and correspondent camera frame axis, when the object frame origin transformed to the camera frame origin. The relationship between any point $X_o$ in $\{O\}$ and the correspondence point $X_c$ in $\{C\}$ can be expressed as follow: 

\begin{equation}
\begin{array}{c}
\mathbf{X}_{\mathbf{c}}=[R \mid T] \mathbf{X}_{\mathbf{o}} \\

z_c\left[\begin{array}{l}
u \\
v \\
1
\end{array}\right]=K\left[\begin{array}{ll}
R & T
\end{array}\right]\left[\begin{array}{c}
x_o \\
y_o \\
z_o \\
1
\end{array}\right]
\end{array}
\end{equation}

Where $R$ is:
$$R=R_{z}(\alpha) R_{y}(\beta) R_{x}(\gamma)$$

\resizebox{0.95\columnwidth}{!}{%
$=\left[\begin{array}{ccc}
\cos \alpha & -\sin \alpha & 0 \\
\sin \alpha & \cos \alpha & 0 \\
0 & 0 & 1
\end{array}\right]\left[\begin{array}{ccc}
\cos \beta & 0 & \sin \beta \\
0 & 1 & 0 \\
-\sin \beta & 0 & \cos \beta
\end{array}\right]\left[\begin{array}{ccc}
1 & 0 & 0 \\
0 & \cos \gamma & -\sin \gamma \\
0 & \sin \gamma & \cos \gamma
\end{array}\right]$%
}

\resizebox{1\columnwidth}{!}{%
$= {\left[\begin{array}{cccc}
\cos \alpha \cos \beta & \cos \alpha \sin \beta \sin \gamma-\sin \alpha \cos \gamma & \cos \alpha \sin \beta \cos \gamma+\sin \alpha \sin \gamma \\
\sin \alpha \cos \beta & \sin \alpha \sin \beta \sin \gamma+\cos \alpha \cos \gamma & \sin \alpha \sin \beta \cos \gamma-\cos \alpha \sin \gamma \\
-\sin \beta & \cos \beta \sin \gamma & \cos \beta \cos \gamma
\end{array}\right] }$%
}
\vspace{10pt}


In 3D bounding box definition, in addition to $R$ and $T$ the size of the object is also considered. The size is defined as $D = [d_x,d_y,d_z]$ which is the dimension on each side of the bounding box. Other formulation of 3D bounding box can be defined with the exact eight corners of the 3D bounding box. Using these coordinate and the correspondence in the object coordinate frame, the pose can be recovered.

Earlier, we identified three ways to divide the methods based on input, problem formulation, and whether the method is instance or category level. In the following sections, we consider a fraction of available models to more clearly clarify the area and discuss pros and cons for each method.



\section{Problem Formulation}

Pose estimation problem can be formulated as direct classification \cite{nejatishahidin2022object}, regression \cite{Amini2021T6DDirectTF}, 2D-3D correspondences \cite{Peng2019PVNetPV}, or 3D-3D correspondences \cite{Zhao2021CORSAIRCO}. The following sections discuss various methods, along with some of the common loss functions associated with them. 

\subsection{Classification}

\begin{figure}[htbp]
\centerline{\includegraphics[width=\linewidth]{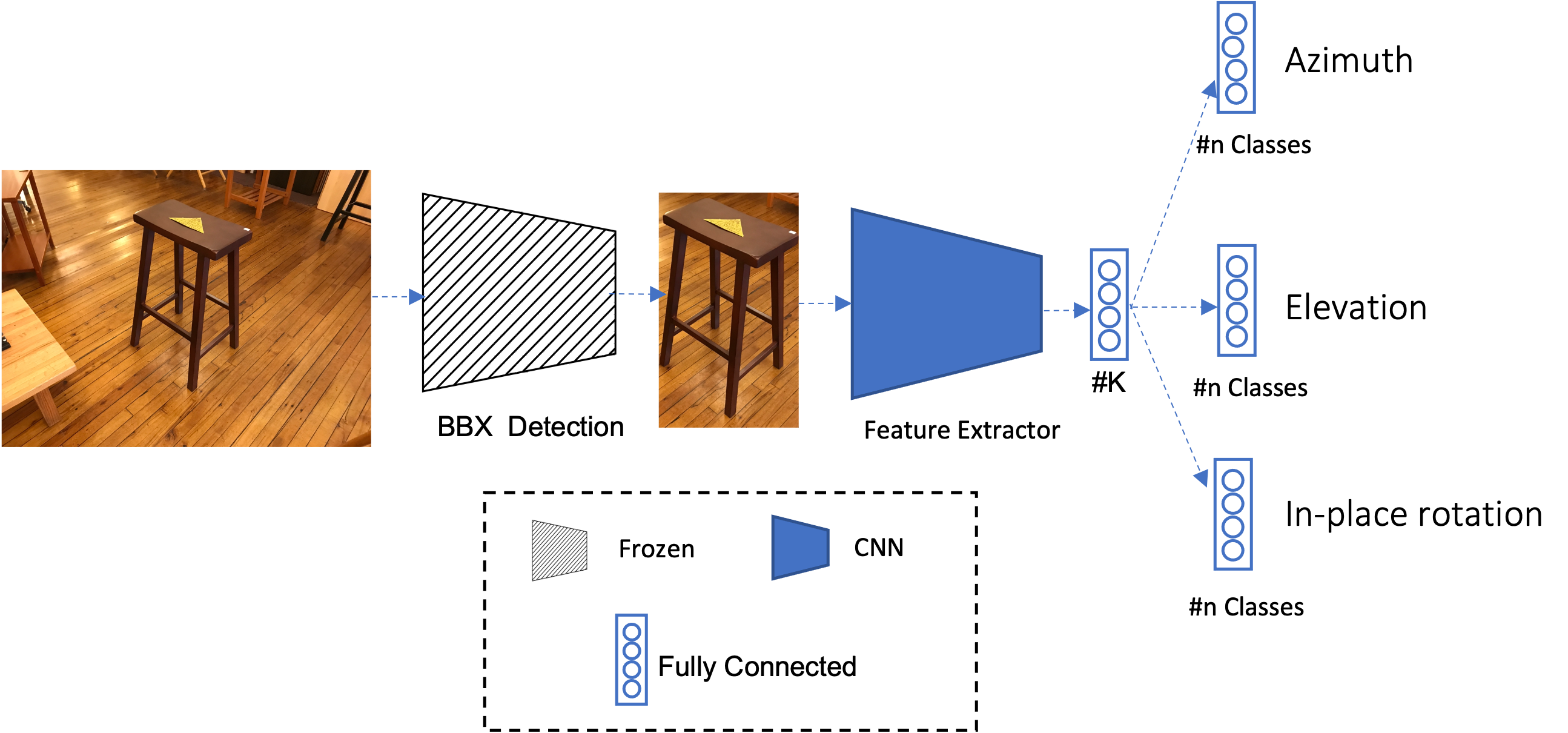}}
\caption{Overall pose classification model. }
\label{ckassification pipline}
\end{figure}

Classification approaches discretize the rotation space \cite{pix3d, nejatishahidin2022object} and cast the 3D rotation estimation into a classification task. Such discretization produces a coarse result, and a post-refinement is essential to get an accurate 6D pose, using either ICP algorithm or regressing the offset \cite{Mousavian20173DBB}. Figure \ref{ckassification pipline} shows a common pipeline. These models consist of Convolution layers for feature extraction followed with MLP at the end to classify the output into a number of bins. The classification loss is the cross-entropy loss as follow: 
$$\mathcal{L}_{cls}=-\sum_{i=1}^{M} y_{o, c_{i}} \log \left(p_{o, c_{i}}\right)$$
Where M is number of classes, $y$ is a binary indicator ( 0 or 1 ) if class label $c$ is the correct classification for observation $o$, and $\mathrm{p}$ is predicted probability of each class $c$.

To get the full 6D pose, the classification is combined with offset regression \cite{Mousavian20173DBB}. The offset can be interpreted as the residual rotation correction that needs to be applied to the center of the bin. This offset can be represented by two numbers, the sine and the cosine of the angle, which are usually predicted with a separate branch of the network. Mostly these models perform better than approaches that directly regress the 6D pose. This results in 3 outputs for each bin $i: ({class\_of\_the\_pose} , cos({offset} ), sin({offset} ))$. The loss function for these methods combines classification and regression losses. The regression part is as follow:
$$\mathcal{L}_{offset}=-\frac{1}{n_{\theta^{*}}} \sum \cos \left(\theta^{*}-c_{i}-\Delta \theta_{i}\right)$$
where offset angle is $\theta^{*}$, $c_{i}$ is the angle of the center of bin $i$ and $\Delta \theta_{i}$ is the change that needs to be applied to the center of bin $i$ (offset).

\subsection{Regression}
\begin{figure}[htbp]
\centerline{\includegraphics[width=\linewidth]{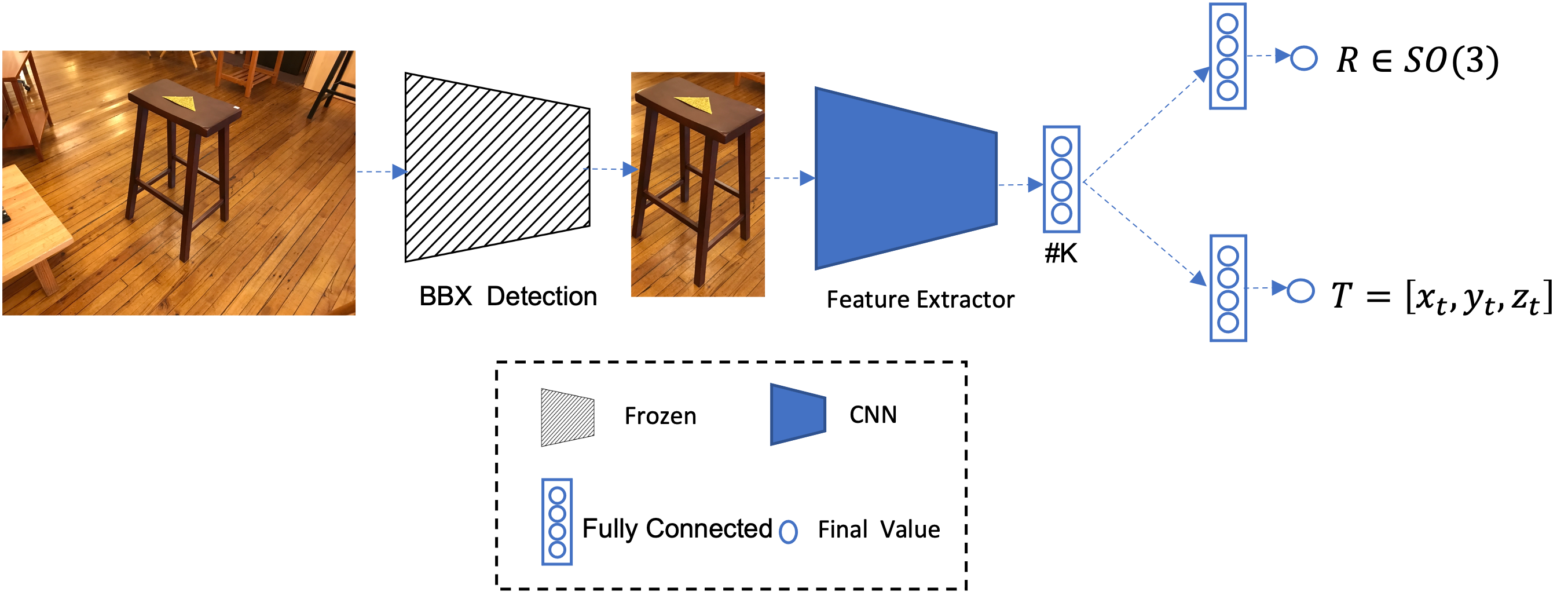}}
\caption{Overall 6D pose regression model. }
\label{regression}
\end{figure}

The pose estimation task can be addressed as a regression problem. The most simple approaches use a Convolutional Neural Network to directly regress the 3D rotation and 3D translation. Figure \ref{regression} shows the vanilla pipeline. These classification methods usually outperform these methods in pose estimation.  Amini et al. \cite{Amini2021T6DDirectTF} regressed the pose with the regression formulation. In this paper the loss is formulated as a distance of the projected points in 2D as follow: 
$$
\mathcal{L}_{\text {pose }}\left(R_{i}, t_{i}, \hat{R}_{\sigma(i)}, \hat{t}_{\sigma(i)}\right)=\mathcal{L}_{R}\left(R_{i}, \hat{R}_{\sigma(i)}\right)+\left\|t_{i}-\hat{t}_{\sigma(i)}\right\|
$$
$$\mathcal{L}_{R}=  \frac{1}{|\mathcal{M}|} \sum_{\mathrm{x} \in \mathcal{M}}\left\|\left(R_{i} \mathrm{x}-\hat{R}_{\sigma(i)} \mathrm{x}\right)\right\| $$

where $\mathcal{M}$ indicates the set of 3D points from provided meshe representation of the object. $R_{i}$ is the ground truth rotation and $t_{i}$ is the ground truth translation. $\hat{R}_{\sigma(i)}$ and $\hat{t}_{\sigma(i)}$ are the predicted rotation and translation, respectively.

\subsection{2D to 3D}
A large body of works in this area address the pose in two stages. They first estimate some correspondences between the image in-camera coordinate and the CAD model in object coordinate space. With image-CAD model pairs of rich textures, traditional feature descriptors such as Scale-Invariant Feature Transform (SIFT) can be used to compute the correspondences. However, these methods cannot address texture-less objects and low-resolution images. In section \ref{2D-3Dsec} we will go into more details about how to find these correspondences. Second, the correspondences are used along with Perspective-n-Point (PnP) algorithm and Random sample consensus (RANSAC) to calculate the 6D pose. PnP solves for the pose given a set of 3D points in the world coordinate frame, their corresponding 2D projections in the image, and a calibrated camera. The PnP can be solved with three points correspondences in the minimal form, called P3P. This results in four real, geometrically feasible solutions. Therefore, a fourth correspondence can be used to remove ambiguity. A more recent method, Efficient PnP (EPnP), is a method that solves the general problem of PnP for more than four correspondences. 

PnP suffers from outliers in the set of point correspondences. RANSAC can be used in conjunction with PnP to make the final solution for the camera pose more robust to outliers. RANSAC is an iterative algorithm that chooses a random subset of the original correspondences (or data points). The PnP is computed for that set (or other mathematical models for different tasks). All other data are then tested against the computed pose (or model). According to some model-specific loss functions, points that fit the estimated model well are considered part of the consensus set. The estimated model is reasonably good if most points are in the consensus set. Finally, the model will improve by re-estimating it using all members of the consensus set. Compared to classification and regression, this method is more informative and robust to occlusions and truncation.


 
\subsection{3D to 3D}
In the presence of the 3D data, 6D pose estimation can be addressed using correspondences between the 3D representation of the object and the CAD model. 
Later these correspondences are used to solve the translation and rotation. In order to calculate translation, we only need to subtract the average of the point in one coordinate system with the average of their correspondences in the other coordinate system. If the set of points in coordinate frame one is $A$ and the correspondences in coordinate frame two is $B$, and the single value decomposition of $BA^T = USV^T$, then the $R= VU^T$.

To address the 6D pose and dimensions of unseen object instances in an RGB-D image, group of work try to combine the regression and correspondences formulation. These methods regress the 3D correspondence of every pixel of points inside the object's mask to its normalized CAD model. NOCS \cite{Wang2019NormalizedOC} presented the Normalized coordinate space, a shared canonical representation for all possible object instances within a category, to handle different and unseen object instances within the category. The region-based neural network, based on mask\_RCNN \cite{He2020MaskR}, is then trained to directly regress the correspondence from observed pixels to this shared object representation (NOCS) along with other object information such as class label and instance mask. These predictions can be combined with the depth map to jointly estimate the 6D pose using Umeyama \cite{Umeyama1991LeastSquaresEO} and RANSAC \cite{Fischler1981RandomSC}. One of the challenges in NOCS is requiring CAD models for all categories in the training stage. DRACO \cite{DBLP:journals/corr/abs-2011-12912} proposed a self-supervised version of the NOCS, which doesn't require NOCS map annotation anymore. These methods work for the category of objects with slight appearance differences such as bowl, cup, laptop, and can. 

\section{Input }
Pose estimation methods can be classified according to their input modalities. Representations used in different techniques can have a significant impact on their effectiveness and the properties of the model. Various representations are investigated for current deep learning models in \cite{deeprepleran}. For pose estimation tasks, inputs to models can be RGB images or 3D information. 
The 3D input uses 3D information solely or along with the RGB image. 3D data can be represented as RGB-D images, CAD models, PointClouds, Voxels, Meshes, or Truncated Sign Distance Fields (TSDF). These representations will be discussed in more detail in Section \ref{3D-input}.

\subsection{RGB Images} 
Using only RGB images to estimate the pose is a challenging task. Inherently, the shape and geometry of the object give away its pose regardless of its appearance. In order to pose estimation models capture this information from pixel-level RGB images, they require a huge number of training data. However, the labeling process is usually challenging, costly, and time-consuming. For example, in the Pix3D dataset \cite{pix3d}, the CAD models for all the objects are collected first. Then, the corresponding Key points between the CAD models and RGB images are hand-labeled. The pose calculated based on these correspondences using the Efficient Perspective-n-Point (EPnP) algorithm \cite{Lepetit2008EPnPAA}. Despite all the challenges, the availability of these datasets, \cite{pix3d, xiang_wacv14, hinterstoisser2012accv, Brachmann2014Learning6O, Hodan2018BOPBF}, and easy access to RGB images in the real-world make this area an active research problem. In the following, we describe RGB-based approaches and divide them based on the commonly used methods.

\subsubsection{Direct Classification or Regression}
As discussed earlier, one way to estimate the pose is classification and regression models applied on top of RGB images. It either the directly fed to the model or a detected bounding box of the object is fed to the model, and then the pose will be estimated. We call these single shot and two-stage models. 

\textbf{Two-Stage.} The early approaches used two stages network. The models first detect the 2D object bounding box \cite{Ren2015FasterRT} and then estimate a single object pose from it. In Viewpoint and Keypoint \cite{Tulsiani2015ViewpointsAK}, pose estimation addressed as a classification problem from the 2D bounding box. The pre-trained VGG \cite{Simonyan2015VeryDC} on ImageNet was used to predict the Viewpoint. In Render for CNN \cite{Su2015RenderFC} the same approach applied, but with more than 2 million rendered synthetic images with ground truth pose, resulting in a better accuracy on the pose. Mousavian et al. \cite{Mousavian20173DBB} also used VGG backbone with extra CNN and Fully connected (FC) layers on top to classify the pose from the 2D bounding box and regress the offset. The two-stage model challenge mentioned earlier\cite{Hodan2018ASO} is the inability to estimate the pose for multiple objects in the image at once. Furthermore, using only the bounding box of an object instead of the entire image may result in the loss of useful information. For example, the room layout or object-to-object relations, which are outside of the object's bounding box, can help the model apply some constraints and improve the performance. 

\textbf{Single-Shot.} Later approaches estimate the 2D bounding box and pose of multiple objects simultaneously in a fast single-shot \cite{Poirson2016FastSS} to address mentioned challenges. Poirson et al. \cite{Poirson2016FastSS} network is based on SSD \cite{Liu2016SSDSS}. These approaches are all addressed pose estimation as a classification problem using only RGB images. The SSD-6D \cite{Kehl2017SSD6DMR}, which is also an extension to SSD, uses multi-scale features to regress the objects bounding boxes and classify the pose to the discrete number of viewpoints. It later applies a refinement stage using an edge-based algorithm to compute the exact pose. These methods are faster than the two-stage models, but they need many training samples. A single network both needs to address the category agnostic pose and 2D bounding box from RGB pixel-level information.

\subsubsection{With CAD Models}
Having the CAD model of objects is a big advantage for pose estimation as the object's geometry is responsible for the pose rather than the appearance. In order to get the benefit of the CAD models, RGB-based methods use them in the training stage to render training samples. They can also be used in training and testing as an input to model along with the image. It's hard to apply the second group to the real-world data, as the CAD models of all instances are not available. 

\textbf{In Training Stage.} A group of RGB-based approaches is trying to take advantage of the CAD model during the training stage. With textured CAD models, one way to use them is to augment training samples. In \cite{Wu2018RealTimeOP} the author used these rendered synthetic data and object masks as an intermediate representation. The model consists of 2 parts, one segmentation and the other one using the mask to classify the pose. These models mostly fail when the mask's quality is poor or when the object is occluded or truncated. 

\textbf{In Training and Testing.} Rendered views of textured CAD models can be used during testing as well. SilhoNet \cite{Billings2019SilhoNetAR} used the renderings to improve the quality of the mask, especially for occluded objects. It predicts the unoccluded mask of the occluded objects and then classifies the pose. A better mask will be estimated using rendered RGB images of textured CAD models from different viewpoints and feeding them along with the original object bounding box to the model. This mask will be fed to a model to classify the pose. Although this method addresses the occlusion problem, having the textured CAD models for each instance is challenging. In these models, objects' masks are used as an intermediate representation to bridge synthetic and real data, simplifying the transfer of models trained on synthetic data to the real-world.

\subsubsection{2D-3D Correspondences}
\label{2D-3Dsec}
Estimating correspondences between the image in camera coordinate and the object coordinate space is a commonly used approach. Later, these correspondences are used along with the PNP algorithm to calculate the pose. These correspondences can be the 2D projection of Corners of the 3D Bounding Box or keypoints.

\begin{figure}[htbp]
\centerline{\includegraphics[width=\linewidth]{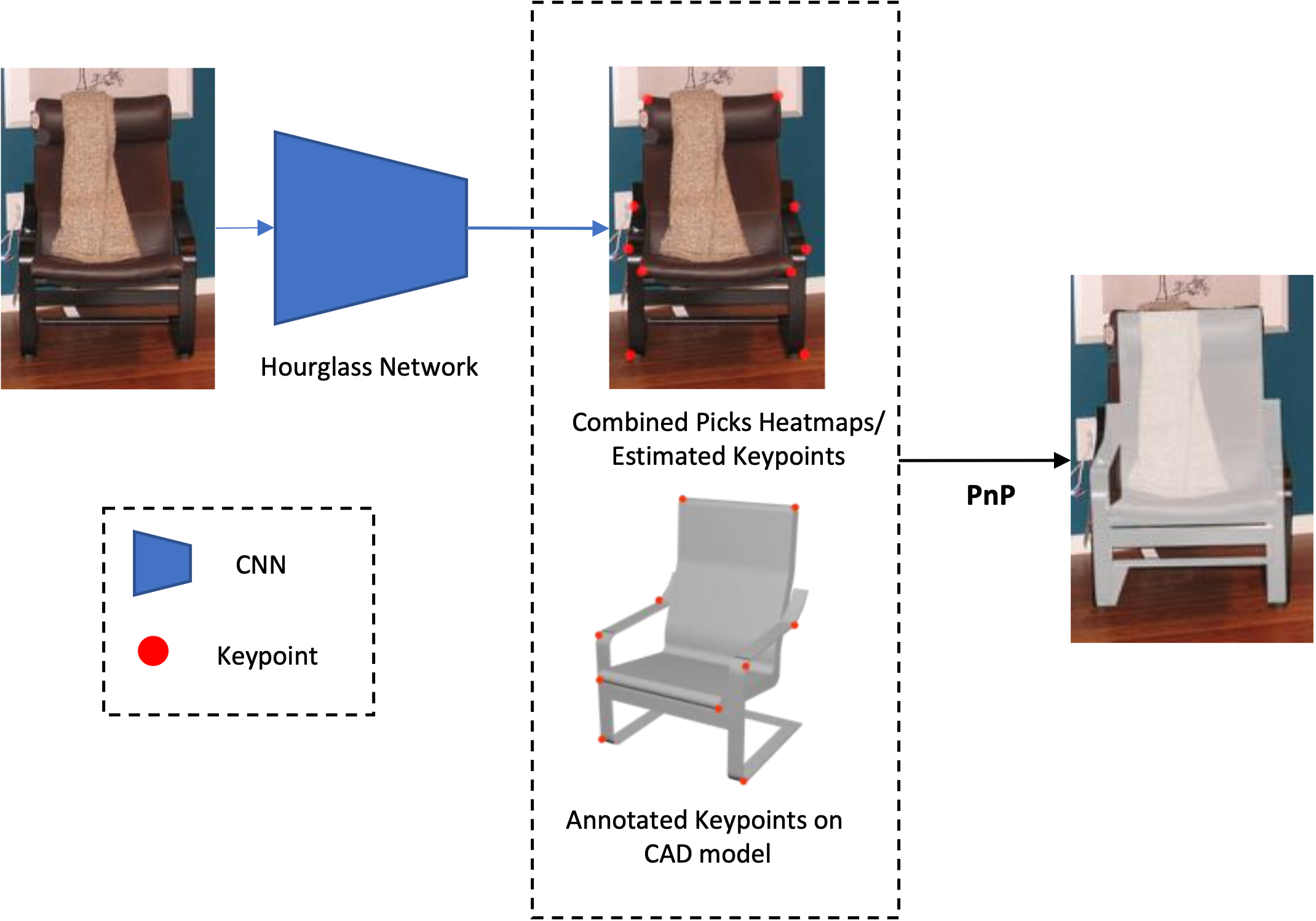}}
\caption{Schematic of keypoint-based 6D pose estimation, adopted from \cite{pix3d}.}
\label{2D-3D}
\end{figure} 

\textbf{Keypoint-Based.} These methods adopt a two-stage pipeline, they first predict 2D keypoints of the object and then compute the pose. For objects of rich textures, traditional methods \cite{Lowe1999ObjectRF} detect local keypoints robustly, so the object pose is estimated both efficiently and accurately, even under cluttered scenes and severe occlusions. However, traditional methods have difficulty handling texture-less objects and processing low-resolution images. Recent works define a set of semantic keypoints and use CNNs as keypoint detectors to solve this problem. In Semantic keypoint \cite{DBLP:journals/corr/PavlakosZCDD17} the keypoints are predicted using hourglass \cite{Newell2016StackedHN} network. Given the predicted 2D keypoints and their correspondences on the 3D model, one naive approach simply applies an existing PnP algorithm to solve the 6-DoF pose. Due to occlusions, false detection in the background, and unavailability of exact 3D CAD models, the 6D pose is solved as an optimization problem. A simple pipline for estimating correspondences is shown in \ref{2D-3D}. The correspondences would be the estimated keypoints and annotated keypoints on CAD models. The $L_2$ Loss is minimized during training to estimate the right keypoints as follow:
$$ \mathcal{L}_2=\sum_{i=1}^{n}\left(k_{\text {true }_i}-k_{\text {predicted }_i}\right)^{2}$$

\textbf{Corners of Bounding Box.} In BB8 \cite{Rad2017BB8AS} the correspondences are in the form of 2D projections of the eight corners of the object 3D bounding boxes. 3D pose can then be estimated using PNP algorithm and the corners correspondences. 
\textbf{Voting-Based Models.} To address the occlusion problem for keypoint detection, PVNet \cite{Peng2019PVNetPV} predicts unit vectors pointing to keypoints for each pixel in the mask of the object and localize 2D keypoints in a RANSAC voting scheme. Later, the pose will be calculated using the PnP algorithm and predicted 2D keypoints and 3D keypoints.

\subsubsection{Transformer-Based Models.} 

The recently proposed language model, Transformer \cite{Vaswani2017AttentionIA}, attracts lots of attention in all areas. T6D-Direct \cite{Amini2021T6DDirectTF}, inspiring from Transformer-based object detection model DETR \cite{Carion2020EndtoEndOD}, proposed an approach which formulate 6D object pose direct regression as a set prediction problem. It is a single-stage direct method that directly estimates multi-object poses and bounding boxes. It predicts the center, height, width of each bounding box, and an extra branch to regress the pose.

\subsubsection{2.5D Sketch.} As mentioned earlier, training a model for learning geometry information directly from the pixel level RGB images requires a large amount of training data and costly pose annotations. The resulting models do not generalize well even for the same instance with different textures or backgrounds, as they are more prone to RGB image pixels. This makes transferring from one domain (synthetic) to another domain (real-world) hard. 
The goal of the 2.5D sketch estimation step is to distill intrinsic object properties from input images, such as geometry structure, while discarding properties that are non-essential for pose estimation, such as object texture. These models are more likely to transfer from synthetic to real data, as they are less sensitive to pixel-level information and more dependent on the intrinsic structure. In addition, they can learn more effectively from synthetic data. This is due to rendering realistic 2.5D sketches without modeling object appearance variations in real images, including lighting, texture, etc., are applicable. However, applying occlusion and some real-world challenges are still problematic. 

\textbf{CAD Model Supervision in Training.} Pix3D \cite{pix3d} trained a model to estimate mid-level representations (surface normal, silhouette, depth) on Pix3D dataset using a CAD model supervision during training. Later, these mid-level representations are fed to a CNN model to classify the pose and reconstruct the 3D Voxel representation. 

\textbf{No CAD Model Supervision.} A recently published paper \cite{nejatishahidin2022object} generated these mid-level features without the CAD model supervision during training. It proposed training a lightweight CNN network on top of Taskonomy \cite{Zamir2018TaskonomyDT} generic mid-level features. The results show that these representations help the model outperform available models by a considerable margin with a small amount of data.

\begin{figure*}[htbp]
\centerline{\includegraphics[width=\linewidth]{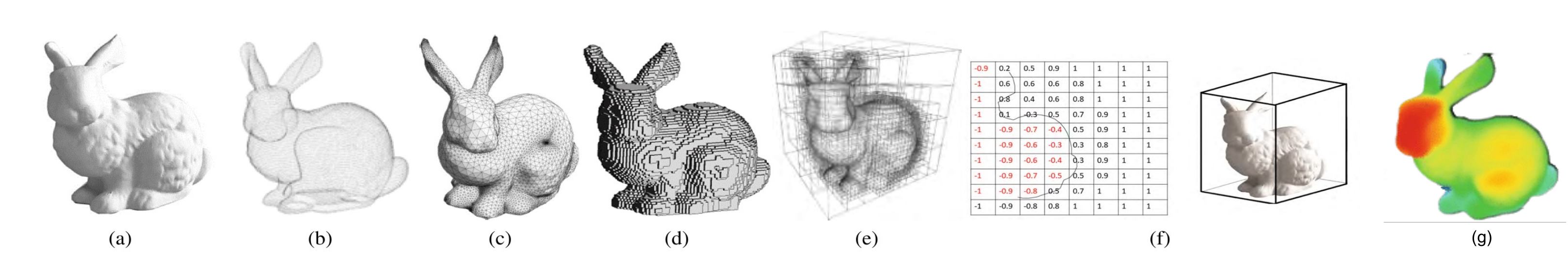}}
\caption{The visualization of different 3D representations for Stanford bunny \cite{Naseer2019IndoorSU}. (a) 3D CAD model. (b) Pointcloud. (c) Mesh. (d) Voxel. (e) Octree. (f) TSDF. (g) Depth map.}
\label{voxel, mesh pointcloud, cad model}
\end{figure*} 

\subsection{3D Input}
For pose estimation models, 3D data provide considerable geometric and structural information. A 3D translation, for example, is the depth of the object center. However, determining the center of an object in a partial pointcloud is a challenging task. The 3D data can be represented in different forms.

\subsubsection{3D Data Representations}  
\label{3D-input}
The 3D input data can be represented as CAD model, 3D Point-cloud, Mesh, Voxel, Octree, Truncated Signed Distance Field (TSDF), or depth-map. Figure \ref{voxel, mesh pointcloud, cad model} shows these representations. A pointcloud is an unordered set of points in three dimensions. Points are spatially defined by the $X, Y, Z$ coordinates. This representation is memory efficient and can be easily converted to any other representation. A voxel can be seen as a 3D base cubical unit that can be used to represent 3D models, which require large amounts of memory. Meshes represent the surface of 3D data with polygons. They are particularly used in computer graphics to represent surfaces or in modeling to discretize a continuous surface. An octree is a voxelized representation of a 3D shape that provides high compactness. Its underlying data structure is a tree. A TSDF is a 3D voxel array representing objects within a volume of space in which each voxel is labeled with the distance to the nearest surface. A depth map is an image or an “image channel” containing information related to the distance of the points constituting the scene from the camera coordinate frame. The RGB-D representation attaches the depth map and color information (RGB).

\subsubsection{Pointcloud Registration} 
Pointcloud registration algorithms similar to Iterative closest point (ICP) has been used for a long time to estimate the pose of 3D shapes \cite{Besl1992AMF}. The 6D pose can be calculated using ICP from CAD models and the object depth. However, these algorithms usually fail when only a part of pointcloud is visible, or the rotation difference is significant. The Deep Closest Point \cite{Wang2019DeepCP} instead of using the points value, learned representations for pointcloud registration. This method can address significant rotation in comparison to other pointcloud registration models \cite{Yang2016GoICPAG, Zhou2016FastGR, Aoki2019PointNetLKR}. Some other groups of papers also use ICP after pose classification step as a refinement stage \cite{Kehl2017SSD6DMR}. 

\subsubsection{3D Bounding Box Detection} 
To detect the 3D bounding box, Deep Sliding Shapes \cite{DeepSlidingShapes} Use a 3D Voxel representation of the depth as input. A fully convolutional 3D network extracts 3D proposals with different receptive fields at different scales. Later they use the 2D RGB features and these 3D proposals to regress the 3D bonding box and classify its category. In \cite{8237757} they used RGB and pointcloud representation. The 2D image is used to reduce the search space in 3D. They use the 2D objects detection from RGB to crop the 3D scene into a frustum. Surface normal is used to estimate the object orientation within the frustum. A multi-layer perceptron is used to regress the object boundaries. This method fails when the bounding box doesn't perform well, or only part of the object is visible. 

\subsubsection{Keypoint-Based Techniques} 

Like keypoint estimation in RGB images, the keypoints can be estimated from RGB and depth. As mentioned earlier, the keypoint-based models are sensitive to occlusion. These methods are primarily instance-level \cite{Liu20203DPVNetP3} and require the exact keypoint annotations from the exact CAD models. In addition, the ground truth label for keypoints on RGB image and CAD models are required. Georgakis et al. \cite{Georgakis_2018_CVPR} proposed an approach, which is Category-level, and nonsensitive. In addition, the keypoint annotation are not required. The model learns consistent keypoint selection across different modalities with the pose supervision, the local features are enforced to be viewpoint invariant and modality invariant. This paper rendered different views of CAD models and forced the similar points from different view result in the same feature descriptor even for the RGB image.



\subsubsection{Voting-Based Techniques} 
Another group of approaches addressed the occlusion with voting. VoteNet \cite{Ding2019VoteNetAD} proposed an approach for the whole scene, represented as a pointcloud. The points were down-sampled; each point then voted for its object's centroid in the scene. The 3D bounding box corners are regressed on the top of clustered votes. VoteNet stated that by concatenating RGB and depth features, the performance does not improve . ImVoteNet\cite{DBLP:journals/corr/abs-2001-10692} proposed an alternative to VoteNet so the RGB image can be used effectively along with the pointcloud. The author used the image to reduce the search space for the object's center. Also, the color texture in the image used to provide a strong semantic prior. 3DPVNet \cite{Liu20203DPVNetP3}, inspired from VoteNet \cite{Ding2019VoteNetAD} and pvnet \cite{Peng2019PVNetPV}, employed deep learning and Hough voting simultaneously to achieve a patch-level 3D Hough voting method for object 6D pose estimation.

\subsubsection{CAD Model and Pose} 

CAD models bring significant geometry and structure priors to the models. \cite{Kuo2020Mask2CAD3S, Kuo2021Patch2CADPE}  learn a global descriptor either from the RGB image to retrieve the most similar CAD model from a database of available CAD models. Mask2CAD \cite{Kuo2020Mask2CAD3S} approach learns to map the RGB image and the renderings of the CAD model to a shared embedding space. The model was trained via contrastive loss between positive and negative pairs of the image-CAD. The model estimates the pose as a classification problem with regressing the offset. Pathch2CAD\cite{ Kuo2021Patch2CADPE} is based on Mask2CAD with a focus on addressing occlusion and new viewpoints challenges. They learned a shared image-CAD embedding space by embedding patches of detected objects from the RGB images and patches of CAD models. By establishing patch-wise correspondence between image and CAD, they can establish object correspondence based on part similarities, enabling more effective shape retrieval for new views and occluded objects. 

\subsubsection{3D-3D Correspondences} 
\begin{figure}[htbp]
\centerline{\includegraphics[width=\linewidth]{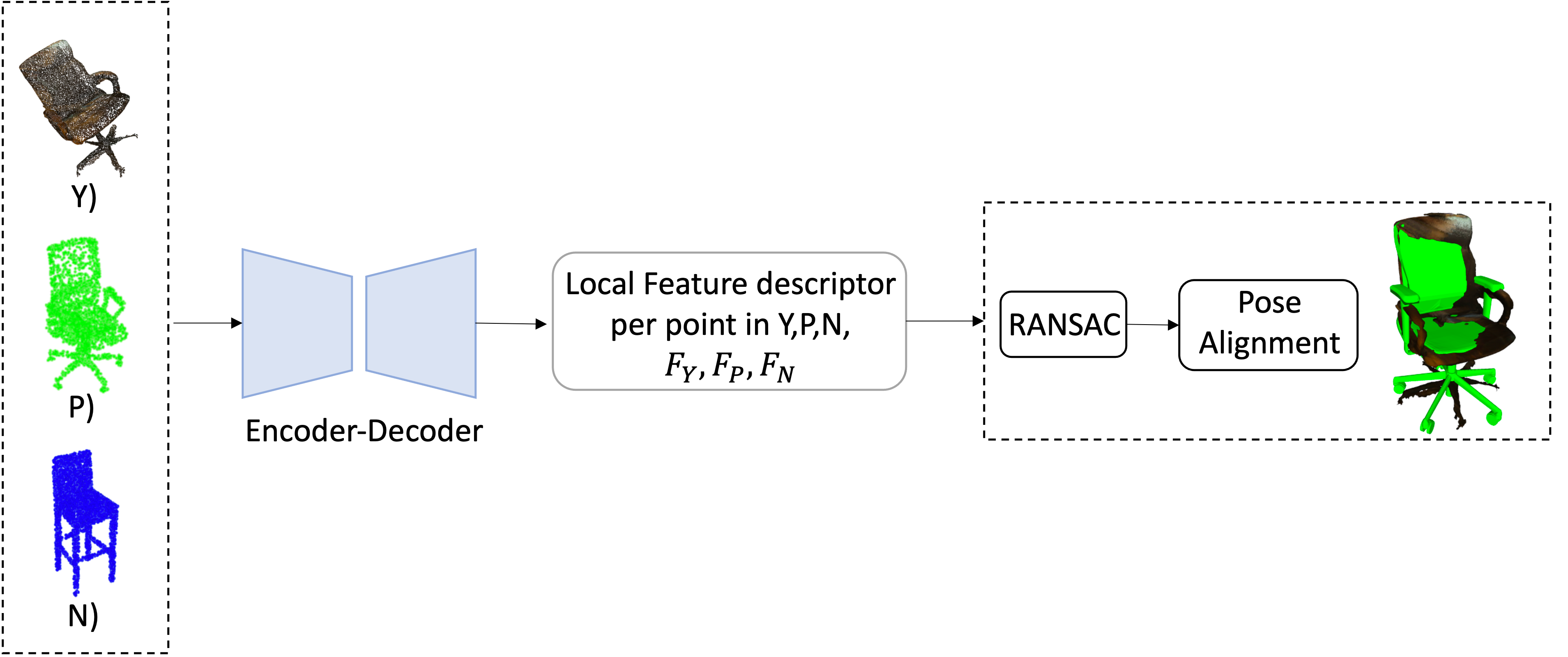}}
\caption{Schematic of 3D-3D correspondence-based 6D pose estimation, similar to \cite{Zhao2021CORSAIRCO}. (Y) is object partial pointcloud representation in camera coordinate frame. (P) is a full pointcloud representation of ground-truth CAD model in object coordinate frame. (N) is a full pointcloud representation of other CAD model in object coordinate frame. The Encoder-Decoder architecture can be either based on sparse convolutions \cite{FCGF2019} or Pointnet++ \cite{DBLP:journals/corr/QiYSG17}.}
\label{3D-3D}
\end{figure}

In order to address the pose we can learn 3D to 3D correspondences and then solve the transformation. The 3D input data can be represented with pointcloud and its corresponding positive and negative pairs (pointclouds of CAD models). Using contrastive loss, an Encoder-Decoder model can be trained to generate local features per point for all the inputs. The features of the corresponding points should be similar and far from the others. During testing, Local features are generated for the CAD model and the object pointcloud. Then matched pairs are used to recover the pose of the query using RANSAC. Figure \ref{3D-3D} show this pipeline. 
The contrastive loss for feature learning can be represented as follow: 
$$\begin{aligned}\mathcal{ L}_{\mathrm{con}}\left(\mathbf{F}^{\mathbf{X}}, \mathbf{F}^{\mathbf{Y}}\right) &=\sum_{(i, j) \in \mathcal{P}_{\mathcal{L}}} \max \left(0,\left\|\mathbf{f}_{i}^{\mathbf{X}}-\mathbf{f}_{j}^{\mathbf{y}}\right\|_{2}-p_{+}\right)^{2} \\ &+\sum_{(i, j) \in \mathcal{N}_{\mathcal{L}}} \max \left(0, p_{-}-\left\|\mathbf{f}_{i}^{\mathbf{x}}-\mathbf{f}_{j}^{\mathbf{y}}\right\|_{2}\right)^{2} \end{aligned}$$
Where $\mathcal{N}_{\mathcal{L}}$is the positive pairs of points and $\mathcal{N}_{\mathcal{L}}$ is the negative pairs of points. $\mathbf{F}^{\mathbf{X}}$ and $\mathbf{F}^{\mathbf{Y}}$  are associated features to two pointclouds from the training set. $p_{+}$ and $p_{-}$ are the positive and negative thresholds.

In CORSAIR \cite{ Zhao2021CORSAIRCO} along with learning the point-wise local features to address the correspondences, a retrieval block is trained to generate a global shape embedding. During testing, given a pointcloud, using the pre-computed CAD models global embedding, the nearest neighbors model is retrieved. Local features are then generated for both pointclouds, and matching pairs are used to recover the pose of the query using RANSAC. 

\section{Instance vs Category Level Approaches }
There is a large body of works focusing on instance-level 6D pose estimation. These models mainly address the pose for the top of the table objects (e.g. cup, laptop, camera, and can). The dominant techniques in these models are trying to do matching between either the RGB or RGB-D input, and the CAD model \cite{DBLP:journals/corr/abs-2111-13489,labbe2020,DBLP:journals/corr/abs-1908-07433} but having the CAD model for all instances of all categories is not feasible in real-world. The challenges mainly encountered at the level of instances are viewpoint variability, texture-less objects, occlusion, and clutter. Instance level approaches fail when it comes to the unseen instances within the seen categories. 
In order to models robustly work in a generalized fashion NOCS \cite{Wang2019NormalizedOC} proposed an approach for category level pose estimation for the top of the table objects with a dataset. 
Most of the category-level pose estimation models \cite{nejatishahidin2022object,pix3d,Mousavian20173DBB,Georgakis_2018_CVPR} works for main categories of objects (e.g., chair, sofa, desk, table, bench, and car). These models mostly make simplifying assumptions about the pose using available constraints in the real-world. For example, the categories mentioned above are parallel to the ground, and only azimuth and elevation are estimated. The main challenges of the category-level 6D object pose estimation are intra-class variation and unseen instances in the test stage.

\section{Datasets}
Datasets can be divided into two main groups: Top of the table objects, which are addressed mainly by the instance-level approaches, and Main categories of objects, which mostly address the category-level approaches. 

\subsection{Instance-Level Datasets}
Linemod \cite{ hinterstoisser2012accv}, Linemod-Occluded \cite{Brachmann2014Learning6O}, T-LESS \cite{ Hodan2018BOPBF}, IC-MI \cite{Tejani2014LatentClassHF}, IC-BIN \cite{Doumanoglou2016Recovering6O} are the datasets most frequently used to test the performances of instance-level full 6D pose estimators. In a recently proposed benchmark for 6D object pose estimation \cite{hodan2020bop}, these datasets are refined and are presented in a unified format along with three new datasets (Rutgers Amazon Picking Challenge \cite{Rennie2016ADF}, TUD Light, and Toyota Light).

\textbf{Texture-Less Datasets.} Linemod has 15 texture-less household objects with discriminative color, shape, and size. The test sets show an annotated object instance with significant clutter but only mild occlusion.
Linemod-Occluded introduced challenging test cases with various levels of occlusion over Linemod. 
T-LESS contains 30 industry-relevant objects with no significant texture or discriminative color. The objects exhibit symmetries and mutual similarities in shape or size, and a few objects are a composition of other objects. Test images originate from 20 scenes with varying complexity. 

\textbf{Textured Datasets.} IC-MI has four textured households and two texture-less objects. The test images show multiple object instances with clutter and slight occlusion. IC-BIN added test images of two objects from IC-MI, which appear in multiple locations with heavy occlusion in a bin-picking scenario. Rutgers Amazon Picking Challenge contains 14 textured products from the Amazon Picking Challenge, each associated with test images of a cluttered warehouse shelf. TUD Light contains three moving objects under eight lighting conditions. Toyota Light contains 21 objects, each captured in multiple poses on a table-top setup, with four different backgrounds and five different lighting conditions.

\subsection{Category-Level Datasets}
KITTI \cite{Geiger2013VisionMR}, SUN RGB-D \cite{Song2015SUNRA}, NYU-Depth v2 \cite{Silberman:ECCV12}, PASCAL3D \cite{Xiang2014BeyondPA}, and Pix3D \cite{pix3d} are the datasets for main categories of objects including, car, sofa, chair, desk ,table, cabinet, bed, and lamp. 

\textbf{CAD Model is Available.}
In particular, KITTI has three categories, car, pedestrian, and cyclist, with a total number of 80256 labeled objects. It has 14999 images, 7481  for training the detectors, and the remaining is for testing. 
PASCAL3D+\cite{Xiang2014BeyondPA} is a 3D version of  PASCAL VOC 2012 \cite{Everingham10}(airplane, bicycle, boat, bottle, bus, car, chair, dining table, motorbike, sofa, train, and monitor) are augmented with 3D annotations. For each category, more images are added from ImageNet \cite{Russakovsky2015ImageNetLS}, resulting in a total number of 30899 images with annotated objects \cite{Sahin2020ARO}. 
The pix3D \cite{pix3d} dataset has nine categories (bed, chair, table, desk, sofa, wardrobes, bookcase, misc, and tool). For each instance, around 50 images are available; for example, the bed category contains 1000 images with 20 instances (around 50 images per instance). Pix3D contains 10,069 images and 395 CAD models.

\textbf{Depth is Available.}
SUN RGB-D \cite{Song2015SUNRA} has 10335 RGB-D images in total, 3784 of which are captured by Kinect v2 and 1159 by IntelRealSense. The remaining are collected from the datasets of NYU-Depth v2 \cite{Silberman:ECCV12}, B3DO \cite{Janoch2011AC3}, and SUN3D \cite{Song2015SUNRA}. This dataset has bounding box annotation for 10,335 RGB-D images. The dominant categories are chair, table, sofa, desk, bed, cabinet, and lamp. There are around 5K training in the dataset for 37 object categories, and the rest is for testing. The dataset also has room layout annotation, making it a good candidate for a 3D holistic understanding. The NYU-Depth v2 dataset \cite{Silberman:ECCV12} involves 1449 images, 795 of which are of training, and the rest is for the test. Even though it has 894 objects, there are mainly 19 object categories on which the methods are evaluated (bathtub, bed, bookshelf, box, chair, counter, desk, door, dresser, garbage bin, lamp, monitor, nightstand, pillow, sink, sofa, table, tv, toilet). 

\section{3D Scene Modeling}
\label{3D-Total}
\begin{figure}[htbp]
\centerline{\includegraphics[width=\linewidth]{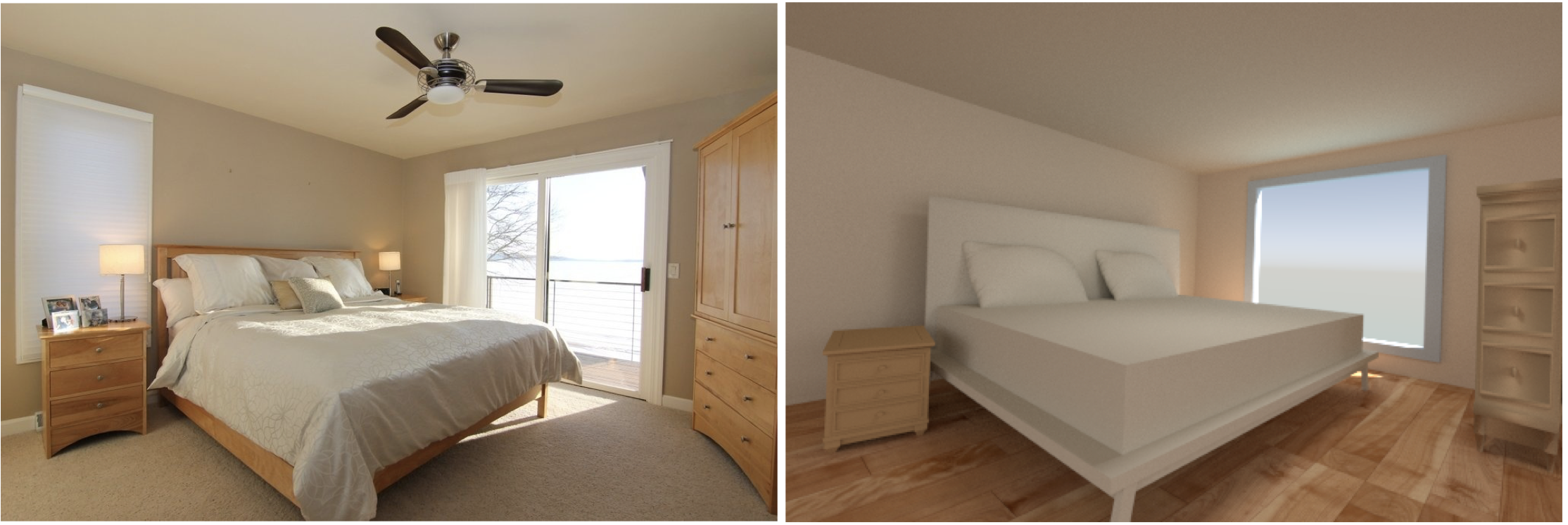}}
\caption{This picture is adopted from \cite{DBLP:journals/corr/IzadiniaSS16}. It shows the idea of fully understanding a 3D scene is composed of object poses, object 3D Reconstruction, camera pose, and layout of the room. }
\label{im2cadimg}
\end{figure}

Understanding 3D scene from a single image is fundamental to various tasks, such as robotics, motion planning, or augmented reality. To fully understand a 3D scene from a single image, we need to know the pose, dimension, and 3D reconstruction of all objects, along with the room layout and the camera pose. With this information, given a single photo of a room, the scene can be synthesized, resulting in a scene that is as close to the photographed scene as possible, see Figure \ref{im2cadimg}. The 3D reconstruction can be represented in any form of CAD model, Voxel, Mesh, or NeRF. 3D Layout estimation aims to reconstruct the intrinsic 3D structure of a room (the geometry of the floor, ceiling, and walls). This task is challenging as the scenes are usually full of various objects and furniture. In an indoor setting, the room layout, object pose, and camera pose are all related and can apply certain constraints to reduce the complexity of the search space. For example, the main indoor objects are always parallel to the ground, which means with knowing the layout, we only have one degree of freedom for the rotation of the object pose (Azimuth). We describe some of the pipelines that address the 3D Visual Understanding problem in the following.






IM2CAD \cite{DBLP:journals/corr/IzadiniaSS16} is a completely automatic system addressing the 3D scene modeling problem. First, the layout of the room is estimated by classifying pixels as being on walls, floor, or ceiling and fitting a box shape to the result. In parallel, they detect all of the main categories of the object such as chairs, tables, sofas, bookshelves, beds, night tables, and windows in the scene using \cite{Ren2015FasterRT}. The CAD model and pose of the objects are estimated by comparing their appearance with renderings of hundreds of beds from many different angles, using a deep convolutional distance metric trained for this purpose. Finally, using the difference between the rendered room and the photograph, they optimize the placement of objects. Although these models perform well, pose estimation and CAD model recovery may not work when the objects are more cluttered and occluded. Additionally, the layout of the room has a box shape, but in reality, it can have another shape. Total3DUnderestanding \cite{DBLP:journals/corr/abs-2002-12212} proposed a network, learning end-to-end for comprehensive 3D scene understanding with mesh reconstruction at the instance level. It is composed of three modules, the Layout Estimation Network, 3D Object Detection Network, and Mesh Generation Network. SceneCAD \cite{Avetisyan2020SceneCADPO} detects the object, estimates the layout and retrieves a similar CAD model. Later on, it builds a graph out of these estimations to understand the holistic of the scene. This enables a robust retrieval and alignment of CAD models to the scan as well as layout generation, resulting in a lightweight CAD-based representation of the scene. In \cite{Dahnert2021Panoptic3S} the author defines 3D scene modeling as a panoptic 3D scene reconstruction. It is composed of predicting the geometry of the scene, semantic labels of all the points, and finding object instances within the image view.

Some of the approaches in this area more focus on using effective representations \cite{Huang2019PerspectiveNet3O, Liu2022TowardsHS} to address the challenges and boost performance. recently proposed approach, InstPIFu \cite{Liu2022TowardsHS}, has used instance-aligned implicit function for detailed object reconstruction. This makes InstPIFu more robust to occlusion and has a better generalization ability on real-world datasets. In addition, it proposes using an implicit representation for the layout of the room instead of the box representation which results in generalization toward all room shapes.

\section{Open Challenges} 
Detecting objects and their 6D poses are an integral part of spatial 3D perception relevant to semantic simultaneous localization and mapping approaches~\cite{SLAM++}, target-driven navigation, autonomous driving, object manipulation, and augmented reality. Although the state-of-the-art deep learning approaches have marked notable advancements by training pose estimation models, the resulting models do not generalize well even to the same instance in different environments, especially when it comes to real-world data. 

The main challenges in pose estimation can be divided into three major categories. 1) Real-world challenges, 2) addressing unseen instances, and 3) addressing unseen object categories. 

It has been noted that model performances are not up to par due to real-world challenges. Figure \ref{challengess} shows some of these challenges. A key factor behind this performance drop is the difficulty in understanding the geometry of an object in real-world settings. For example, understanding the structure of an object which is partially occluded, transparent, or dark is challenging. This type of difficulty is not restricted to only pose estimation; semantic segmentation, object classification, and detection can also fail in these scenarios. Many approaches \cite{nejatishahidin2022object, Mousavian20173DBB, Su2015RenderFC} are dependent on these state-of-the-art object detection \cite{Ren2015FasterRT} or mask segmentation \cite{He2020MaskR} as a part of their pipeline.


\cite{DBLP:journals/corr/IzadiniaSS16, DBLP:journals/corr/abs-2002-12212, DBLP:journals/corr/abs-1810-13049, Yang2022Boosting3O} tries to address some of the real-world challenges for main categories of indoor objects. However, DeMF \cite{Yang2022Boosting3O}, which is the best-performing model on SunRGBD, is $46\%$. For some categories like bookshelves and desks, the performance is even less than $17\%$. Performances also drop when they are applied to other datasets. According to the results, the models have difficulty understanding the geometry when there is noise in the depth data or when occlusion or truncation are present. The promise of voting-based techniques \cite{Peng2019PVNetPV, Liu20203DPVNetP3} are being robust to these challenges. But these models address the occlusion and truncation for a small number of object instances. The other voting-based techniques \cite{Ding2019VoteNetAD, DBLP:journals/corr/abs-2001-10692} vote for the center which improves the translation accuracy. However, these models require the entire scene pointcloud, which is not applicable to many robotics applications depending on pose estimation approaches. 


The second main problem is the difficulty in addressing unseen instances. Most of the RGB-based models, which train an end-to-end CNN model to estimate the pose, do not perform well on other datasets, even on the same categories. As they learn a direct mapping from the pixel level information to the object pose, and they are fitting to the distribution of the dataset, especially when there is no supervision or constraint to learn the geometry information. The models which get the advantage of the 3D information are either using RGB-D or CAD models. Those which are highly dependent on CAD models are mainly limited to the instances in which the CAD models are available. These models are more stable and have a better idea of the object structure. Although some techniques are trying to define a deformable shape priors \cite{DBLP:journals/corr/PavlakosZCDD17} to address the unavailability of CAD models for all the instances, but still not performing well for all the categories, especially the ones with more appearance change within the category. 

Lastly, it is extremely challenging to address novel categories and objects, a newly proposed approach called FS6D \cite{he2022fs6d} inspired by SuperGLUE \cite{DBLP:journals/corr/abs-1905-00537}, a feature matching technique, addressed the object pose estimation as a feature matching problem between few 2D views of exactly the same object  with the ground truth pose annotation and the query image. Although the idea is novel and extends the pose estimation to novel objects in a few-shot manner, still zero-shot techniques for novel object pose estimation are missing. Techniques that understand the geometry and do not highly depend on the texture.




A holistic understanding of the scene can be used to improve pose estimation performance and build more generalized models for addressing real-world challenges in this area. This holistic understanding can be interpreted as object-to-object and object-to-layout relations. These can be applied in the prepossessing step as a constraint \cite{DBLP:journals/corr/abs-1906-02729}, or in the post-processing as an optimization part. 

In addition, multi-view information \cite{Sajnani2021DRACOWS} can help to address truncated or occlusion to some extent. Also, it can assist in applying constraints to the pose of a static object; the pose of a static object depends on the transformation of the camera.

The unavailability of the CAD models for all instances of the objects can be tackled by defining representative shape priors \cite{Tian2020ShapePD} for each category. For ones with more appearance changes, sub-category shape priors can be defined. These shape priors represent CAD models' essential and representative components throughout the entire category. 
\section{Conclusion}
We have presented a short review on state-of-the-art approaches to Object Pose estimation. The approaches are classified according to the essential properties that affect both performance and the targeted objects. However, models continue to fail when they encounter novel objects or challenging scenes. The use of approaches that consider a holistic 3D understanding can assist in addressing challenging real-world scenarios. Another option is to develop models which have a deeper understanding of the object structure.

{\small
\bibliographystyle{ieee}
\bibliography{egbib}
}

\end{document}